\documentclass[sigconf]{acmart}
\usepackage{algorithm}
\usepackage{algorithmic}
\usepackage{newfloat}
\usepackage{listings}
\usepackage{booktabs}
\usepackage{makecell}
\usepackage{xcolor}
\usepackage{hyperref}
\usepackage{bm}

\AtBeginDocument{%
  \providecommand\BibTeX{{%
    \normalfont B\kern-0.5em{\scshape i\kern-0.25em b}\kern-0.8em\TeX}}}

\renewcommand\footnotetextcopyrightpermission[1]{} 

\acmConference[Preprint]{}{}
\begin{document}

\title[DisDepth]{Promoting CNNs with Cross-Architecture Knowledge Distillation\\for Efficient Monocular Depth Estimation}


\author{Zhimeng Zheng}
\affiliation{%
  \institution{Zhejiang University}
  \city{Hangzhou}
  \country{China}}
\email{22110145@zju.edu.cn}

\author{Tao Huang}
\authornote{Corresponding author.}
\affiliation{%
  \institution{The University of Sydney}
  \city{Sydney}
  \country{Australia}}
\email{thua7590@uni.sydney.edu.au}

\author{Gongsheng Li}
\affiliation{%
  \institution{Zhejiang University}
  \city{Hangzhou}
  \country{China}}
\email{22210102@zju.edu.cn}

\author{Zuyi Wang}
\affiliation{%
  \institution{Zhejiang University}
  \city{Hangzhou}
  \country{China}}
\email{zuyiwang@zju.edu.cn}



\begin{abstract}
Recently, the performance of monocular depth estimation (MDE) has been significantly boosted with the integration of transformer models. However, the transformer models are usually computationally-expensive, and their effectiveness in light-weight models are limited compared to convolutions. This limitation hinders their deployment on resource-limited devices. In this paper, we propose a cross-architecture knowledge distillation method for MDE, dubbed DisDepth, to enhance efficient CNN models with the supervision of state-of-the-art transformer models. Concretely, we first build a simple framework of convolution-based MDE, which is then enhanced with a novel local-global convolution module to capture both local and global information in the image. To effectively distill valuable information from the transformer teacher and bridge the gap between convolution and transformer features, we introduce a method to acclimate the teacher with a ghost decoder. The ghost decoder is a copy of the student's decoder, and adapting the teacher with the ghost decoder aligns the features to be student-friendly while preserving their original performance. Furthermore, we propose an attentive knowledge distillation loss that adaptively identifies features valuable for depth estimation. This loss guides the student to focus more on attentive regions, improving its performance. Extensive experiments on KITTI and NYU Depth V2 datasets demonstrate the effectiveness of DisDepth. Our method achieves significant improvements on various efficient backbones, showcasing its potential for efficient monocular depth estimation.
\end{abstract}

\ccsdesc[500]{Computing methodologies~Computer vision}
\ccsdesc[300]{Computing methodologies~Machine learning approaches}

\keywords{Monocular Depth Estimation, Cross-architecture Knowledge Distillation, Computational Efficiency, Local-Global Convolution Module}



\maketitle
\section{Introduction}
Monocular depth estimation(MDE) ~\cite{eigen, alhashim}  has achieved remarkable success by the adventure of automatic feature engineering in deep neural networks. With the advent of better global feature representation in vision transformers~\cite{vaswani}, recent state-of-the-art MDE methods resort to complicated transformer-based backbones~\cite{li,Li_Chen,agarwal} and decoder structures~\cite{bhat,agarwal}, which are computational-expensive and hard to be deployed on resource-constrained devices. For example, PixelFormer~\cite{agarwal} with Swin-L backbone has $\sim$704B FLOPs and $\sim$271M parameters, which is too heavy for edge devices to inference.

\begin{figure}
    \centering
    \includegraphics[width=0.9\linewidth]{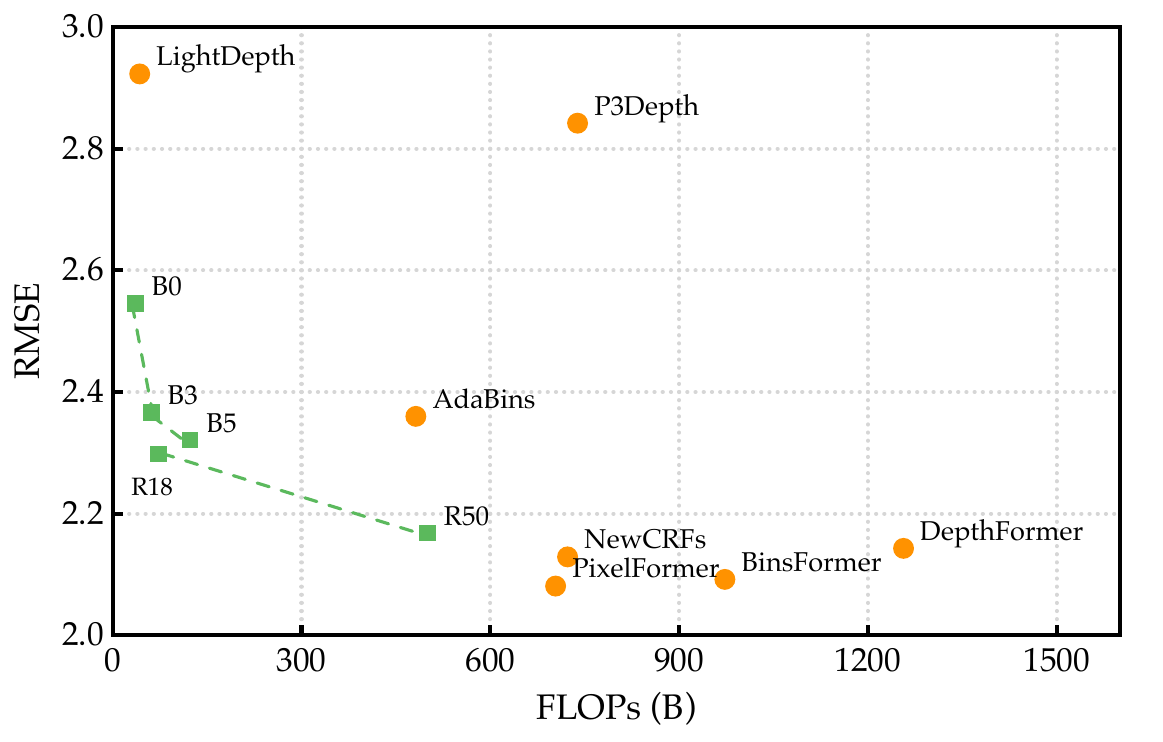}
    \vspace{-2mm}
    \caption{RMSE (lower is better) and FLOPs comparisons of existing MDE models and our DisDepth models on KITTI dataset. DisDepth variants (green squares) obtain competitive performance with significant superiority on efficiency.}
    \label{fig:cmp}
\end{figure}

\begin{figure*}[t]
    \centering
    \includegraphics[width=1\linewidth]{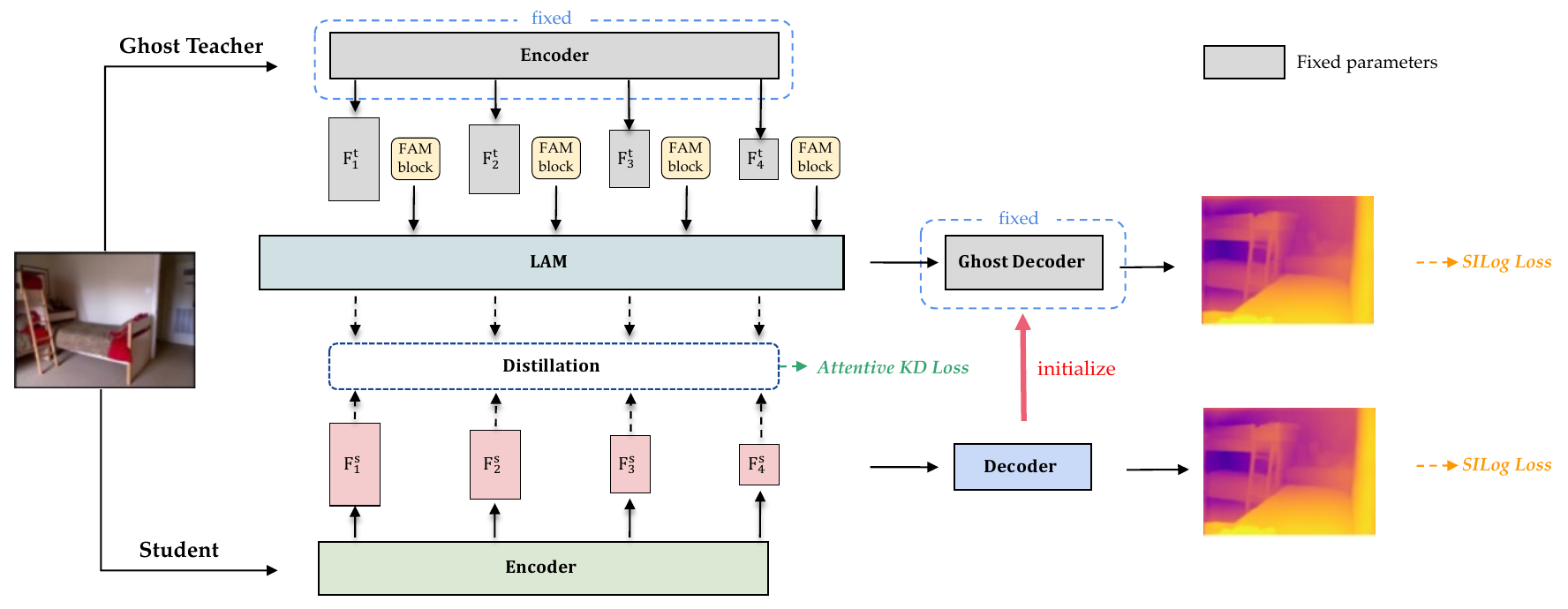}
    \caption{Framework of our cross-architecture KD. We introduce (feature acclimation module) FAMs to adapt the outputs of teacher encoder, followed by a (loss attention module) LAM to identify valuable regions in the features, and these modules are optimized with a ghost decoder and task loss. The distillation is conducted on student features and the adapted teacher features. The parameters of teacher encoder and ghost decoder are fixed.}
    \label{fig:framework}
\end{figure*}

In this paper, we aim to obtain competitive performance while maintaining an efficient inference of MDE model by exploring a CNN-based framework without transformers. This is achieved by proposing (1) an efficient CNN-based MDE framework and (2) a novel cross-architecture knowledge distillation (KD) method that adjusts and distills the transformer features into CNNs. Concretely, as shown in Figure~\ref{fig:framework}, we first construct a simple and efficient CNN-based MDE framework that contains a CNN backbone, a simple decoder with convolutions and upsamplings, and two heads for predicting bin centers and bin probabilities~\cite{bhat}. Then, a local-global convolution (LG-Conv) module is proposed to enhance the capability of global representations in CNNs. Our LG-Conv is inspired by our analysis that the role of self-attention in transformers is to aggregate and broadcast the information globally among tokens, while this global information exchange can also be done with simple convolution and pooling operations. The resulting global branch in LG-Conv is a supplement of the original local convolution, and can be directly utilized on a pretrained backbone without retraining on pretraining dataset. Our LG-Conv is efficient and friendly to deployment, and experiments show that it effectively extracts the global information and boosts the performance.

Besides, we propose a new KD method for cross-architecture distillation from transformers to CNNs. Our basic motivation is to leverage the state-of-the-art transformer-based MDE models to improve our CNN models with KD. However, we empirically find that due to capacity and architecture gaps, the student is difficult to imitate teacher's features, and result in poor distillation performance, and even worse than the CNN teacher with significant worse performance. To understand what restricts the distillation performance, we conduct experiments and find that the devil of cross-architecture KD is in the architecture-intrinsic information\footnote{For a brief, we use a fixed and pretrained student decoder to adapt the teacher features, and the teacher still obtains similar performance, while a significant gain on distillation performance has been obtained.}. Therefore, we intuitively decouple the information in teacher feature into two parts, namely intrinsic information and extrinsic information. The intrinsic information is architecture-independent, and small students with different architectures are arduous in learning this information. While the extrinsic information is crucial for generating the task predictions, and is common and invariant to different architectures; it can be easily adapted and distilled to students.

To this end, we propose to acclimate the transformer teacher with a ghost decoder, which is a copy of the student's decoder, so that we can obtain adapted teacher features that are more appropriate for distillation. Specifically, we inject a feature acclimation module (FAM, a transformer block) into the teacher backbone after each output layer, which is used to acclimate the teacher features. The resulting adapted features are fed into the ghost decoder to get the predictions, which are then passed to the task loss function. This error between prediction and ground-truth is used to optimize the FAM, and other modules including original backbone and ghost decoder are fixed. Through this adaptation, the intrinsic information in teacher features is repressed and task-relevant extrinsic information is acclimated, as the student-trained decoder prefers student-like features. Furthermore, instead of directly optimizing the distillation loss between adapted teacher feature and student feature, we introduce a attentive KD loss, which learns valuable regions in the teacher features, and then use the learned region importances to guide the student to focus more on valuable features.

In sum, our contribution is threefold.
\begin{enumerate}
    \item We propose a simple CNN framework for MDE, with a supplementary local-global convolution module to enhance the global representations. Our framework achieves competitive performance and is more efficient than current state-of-the-art transformer-based frameworks.
    \item We propose a novel cross-architecture knowledge distillation method, which effectively represses the architecture-intrinsic information and adapts task-relevant information in the transformer teacher features for better distillation to CNN student. Additionally, an attentive KD loss is introduced to learn and distill valuable regions in teacher features.
    \item Extensive experiments on KITTI and NYU Depth V2 datasets demonstrate our effectiveness. As shown in Figure~\ref{fig:cmp}, our DisDepth achieves significant efficiency and performance superiorities on various model scales.
\end{enumerate}

\section{Related Work}

\subsection{Monocular Depth Estimation}
Monocular depth estimation (MDE) is an active area of research that aims to regress a dense depth map from a single RGB image. The pioneer, \cite{eigen}, which initiated the trend with its coarse-to-fine network and innovative scale-invariant log loss (SIlog) for optimisation. Building on this foundation, the field embraced architectural advances such as residual networks~\cite{Laina}, multi-scale fusion~\cite{Lee}, and transformers~\cite{bhat, yuan}. On the other hand, some works are proposed to improve the optimization techniques, including reverse Huber loss~\cite{Laina}, classification-based schemes~\cite{Cao}, and ordinal regression~\cite{Fu}. The field has also seen a move towards ordinal regression methods, which are similar to mask-based segmentation in their approach to depth discretisation. By associating each depth value with a confidence mask, the culmination in depth prediction becomes a weighted interplay of these discretised values.

\subsection{Knowledge Distillation}

Knowledge distillation (KD) \cite{Hinton,Chen_Choi,huang2022knowledge,Gupta,zhang2023avatar,huang2024knowledge,zhang2023freekd,Wofk} is an idea seeded in the realm of image recognition by~\cite{Hinton}, it serves as a conduit to transfer knowledge from a high-capacity teacher model to a more compact student model. Its versatility has been affirmed with its myriad applications across various computer vision tasks, including domain adaptation~\cite{Gupta}, object detection~\cite{Chen_Choi}, and learning from noisy labels~\cite{Li_Yang}. In the MDE context,~\cite{Pilzer} and~\cite{Aleotti} spearheaded efforts to harness the power of distillation, setting the stage for subsequent innovations. The application of distillation in depth estimation for resource-constrained environments, particularly mobile systems, is of great significance~\cite{Wang_Lai},~\cite{Poggi},~\cite{Wofk}. The challenge is balancing depth estimation accuracy with computational efficiency, crucial for real-time applications like autonomous driving. This balance becomes even more pronounced in bespoke hardware setups, like dual-pixel sensors~\cite{Garg}. While the journey has begun, the landscape of knowledge distillation in MDE, especially under constraints, remains a dynamic and evolving field.

\section{Our Approach}

In this section, we present our method DisDepth. We begin by discussing the design of our CNN-only MDE framework. We then elaborate on our local-global convolution, which is designed to enhance the global representations of the model. Additionally, we introduce our cross-architecture knowledge distillation approach, where we adapt the transformer teacher with a ghost decoder. Finally, we propose an attentive knowledge distillation loss that helps identify valuable regions for the student model to focus on.

\subsection{Frustratingly Simple CNN Framework}

To improve the performance of MDE, recent state-of-the-art approaches have often relied on complex and heavy decoder structures. For instance, AdaBins~\cite{bhat} proposes a mViT module with multiple transformer layers to enhance the decoded feature pyramid. DepthFormer~\cite{li} introduces an HAHI module that combines deformable cross-attention and self-attention to bridge the encoder and decoder. PixelFormer~\cite{agarwal} utilizes a UNet structure with advanced feature fusion modules, including window cross-attention. However, these sophisticated decoder designs come at the cost of reduced efficiency and deployment challenges. The heavy computational requirements and complexity make it difficult to deploy these models in real-world scenarios.

In our research, we have discovered that by using a powerful backbone encoder, a simple CNN-based decoder can still achieve competitive performance in MDE. This finding has led us to develop a surprisingly simple CNN framework. For instance, in Table~\ref{tab:ghost_teacher_performance}, our framework achieves an RMSE of $2.058$ on the KITTI dataset when using the Swin-Large backbone. In comparison, the PixelFormer model, which employs an advanced decoder, achieves an RMSE of $2.081$. This observation suggests that we can effectively replace complex and advanced decoder approaches with a simpler CNN-based decoder, allowing us to focus primarily on improving the encoder. This insight opens up new possibilities for developing efficient and effective MDE models that strike a balance between performance and simplicity.

\textbf{Implementation details.} Our framework is illustrated in Figure~\ref{fig:framework} (see the bottom-half student model). With an input image, it first encodes its features using an encoder (backbone), then passes the features of the last four stages to get the feature pyramid. A simple convolutional and upsampling decoder~\cite{alhashim} is employed to fuse and extract features from the feature pyramid. The decoder combines information from different scales to generate a final feature map. Following the approach introduced in AdaBins~\cite{bhat}, the feature map is passed through two heads. One head predicts the bin centers, while the other head predicts the bin probabilities. The predicted bin centers and probabilities are further processed to obtain the final predicted depth map.

\begin{figure*}[t]
    \centering
    \includegraphics[width=\linewidth]{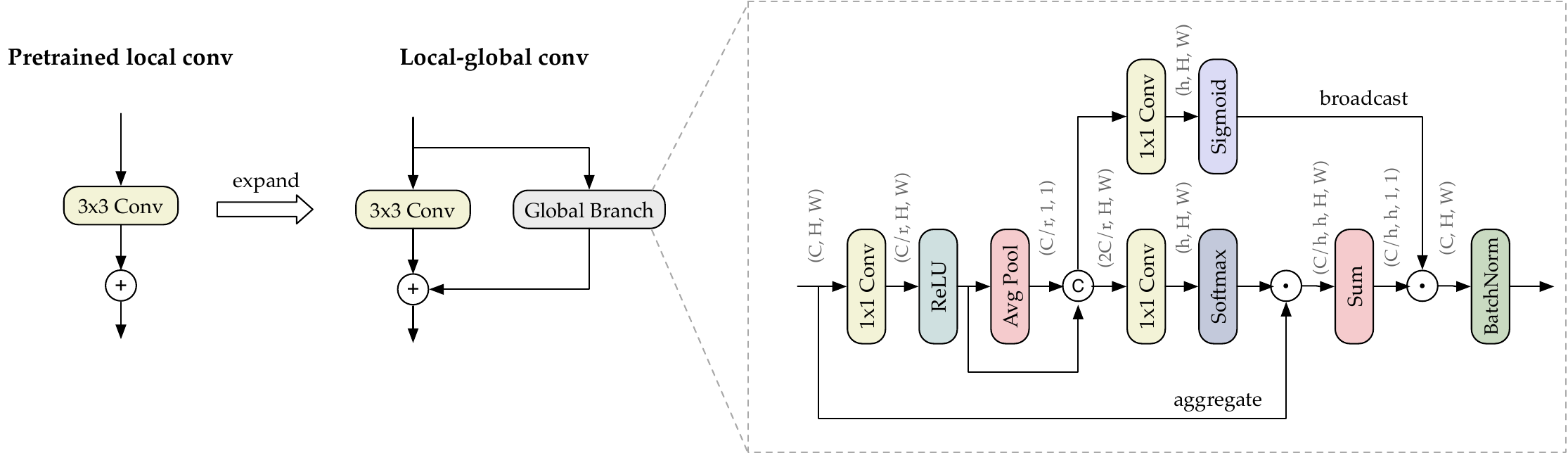}
    \caption{Illustration of our proposed local-global convolution (LG-Conv).}
    \label{fig:global_branch}
\end{figure*}

\subsection{Enhanced CNNs with Local-Global Convolutions}

Compared to CNNs, transformers~\cite{vaswani} are acknowledged to have better global representation capability due to the global self-attention. Some recent methods~\cite{li} have successfully shown that transformers can significantly boost the MDE performance due to better representation power and relationship modeling. However, the computation of self-attention requires has a large peak memory footprint and has not been widely optimized and supported by edge computation frameworks, which have achieved remarkable success by deploying CNNs into real-life productions (see supplementary material for details).

To enhance the CNNs with global dependency, we propose a convolution-based module named local-global convolution (LG-Conv), which works as a supplement of the original local convolution layers. Different from recent approaches in bringing long-term dependency into CNNs, which focus on mixing transformers and convolutions into one block or design complicated computation scheme that cannot be efficiently deployed, our LG-Conv is composed with only convolutions and poolings, and has the following superiorities. (1) Our LG-Conv does not require retraining the backbone. (2) Our LG-Conv is efficient. (3) Our LG-Conv is deployment-friendly.

\textbf{Implementation details.} As shown in Figure~\ref{fig:global_branch}, which a pretrained CNN backbone composed of multiple convolution layers, we expand each $3\times 3$ convolutions with an additional global branch in LG-Conv. The global branch takes the same input feature as the original local convolution. The feature is first transformed using a \textit{Conv-ReLU} structure. Then, the hidden feature is obtained by concatenating the local feature with the global feature, which is obtained through average pooling. We utilize the hidden feature to predict multi-head attentions over the spatial dimension, which are then used to aggregate the pixel features into one feature. Additionally, we introduce a broadcast branch to predict probabilities that determine whether the aggregated global information should be applied to each pixel.
To ensure that the pretrained semantic information is not disturbed by directly supplementing the global branch's feature to the original feature, we propose initializing the scale parameters $\bm{\gamma}$ in the last batch normalization layer with a small value (e.g., $0.001$). This initialization ensures that the global branch has minimal impact on the original features during the initial stages of training, allowing the backbone to evolve smoothly.

\subsection{Acclimating Teacher with A Ghost Decoder}

Among existing methods, the transformer-based models outperform CNN-based models by a large margin. Therefore, to enhance our efficient model, an intuitive idea is to use the state-of-the-art transformer models as the teacher to distill our model (student). However, we empirically find that directly imitating the features of transformer teacher does not gains significant improvements as expected. For instance, as summarized in Table~\ref{tab:kd_vs_ours}, we train our EfficientNet-B0 student with EfficientNet-B5 and Swin-L teachers, and show that, although the Swin-L has an obvious superiority in performance (2.06 RMSE vs. 2.60 RMSE), its distillation performance (2.95 RMSE) is largely behind the EfficientNet-B5 teacher (2.78 RMSE), which has a similar architecture to the student.

To better understand the distinction between CNN features and transformer features, we conduct experiments and make an interesting observation. By using a fixed student-trained decoder to replace the original decoder in the pretrained teacher model, and optimizing the teacher with task loss, we find that the new teacher model achieves similar performance compared to the original teacher. This trial successfully acclimate the teacher features to be similar to the student's features, while still preserving the task-relevant information necessary for good performance.

Based on this observation, we propose a hypothesis that the teacher features can be decoupled into two types of information: architecture intrinsic information and extrinsic information. The architecture intrinsic information is architecture-independent, and the student model finds it challenging to mimic this information due to its smaller capacity and architectural differences. On the other hand, the extrinsic information is task-relevant, and it is common and consistent across different architectures. These features can be easily adapted and imitated by the student model.

Consequently, to achieve effective and student-friendly cross-architecture distillation, we propose acclimating the transformer teacher model with the student's decoder and distilling knowledge from the acclimated features. This process is illustrated in Figure~\ref{fig:framework}. To facilitate this acclimation, we introduce a \textit{ghost decoder}, which is a continuous copy of the student's decoder throughout the entire training period. The original teacher features are adjusted using additional trainable feature acclimation modules (FAMs), the ghost decoder, and task loss, which is a transformer block in the teacher model. Subsequently, we perform distillation between the acclimated features of the teacher model and the features of the student model. The iterative procedure of our DisDepth is summarized in Algorithm~\ref{alg:kd}. Note that the FAMs are trained simultaneously with the student model, at the early of training, the student decoder is not converged and the acclimated features are with low performance, so we add a warmup that starts distillation after a few epochs trained.

\begin{algorithm}[t]
    \caption{Cross-architecture KD with DisDepth}
    \label{alg:kd}
    \footnotesize
    \begin{algorithmic}[1]
        \REQUIRE Student model $\mathcal{S}$, pretrained teacher model $\mathcal{T}$, train dataset $\mathcal{D}_{tr}$, and number of iterations $I$.
        \STATE Integrate FAMs into teacher model
        \FOR{iteration $i$ in $1,...,I$}
            \STATE Get a batch of input $\bm{X}$ and target $\hat{\bm{Y}}$ from $\mathcal{D}_{tr}$
            \STATE Copy the weights of student decoder to ghost decoder
            \STATE Compute the outputs $\bm{F}^{(t)}$ of teacher encoder with $\bm{X}$
            \STATE $\tilde{\bm{F}}^{(t)}\leftarrow \mathrm{LAM}(\mathrm{FAM}(\bm{F}^{(t)}))$
            \STATE Get predictions $\bm{Y}^{(t)}$ with ghost decoder
            \STATE Compute features $\bm{F}^{(s)}$ and predictions $\bm{Y}^{(s)}$ of student
            \STATE Optimize student with $\mathcal{L}_\mathrm{task}(\bm{Y}^{(s)}, \hat{\bm{Y}})$ and $\mathcal{L}_\mathrm{kd}(\bm{F}^{(s)}, \tilde{\bm{F}}^{(t)})$
            \STATE Optimize FAMs and LAMs with $\mathcal{L}_\mathrm{task}(\bm{Y}^{(t)}, \hat{\bm{Y}})$
        \ENDFOR
        \RETURN Trained student model $\mathcal{S}^*$
    \end{algorithmic}
\end{algorithm}

\subsection{Attentive Knowledge Distillation}

\begin{figure}[t]
    \centering
    \includegraphics[width=0.95\linewidth, height=0.82\linewidth]{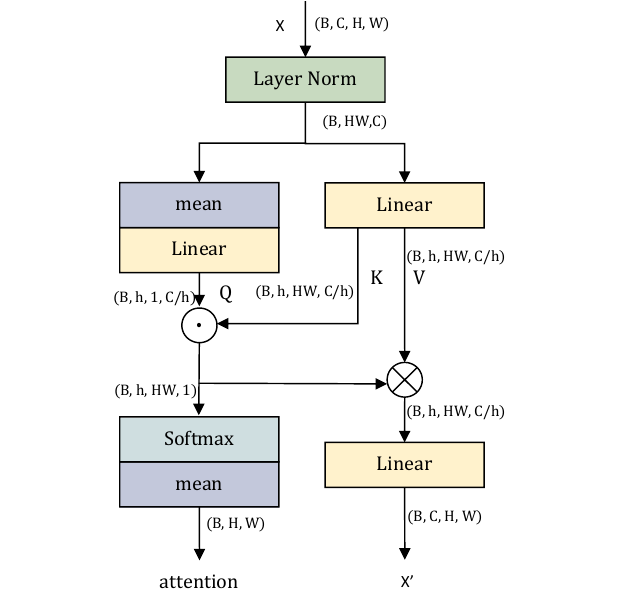}
    \caption{Architecture of loss attention module (LAM).}
    \label{fig:lam}
\end{figure}

We now have successfully built our method of acclimating teacher features. In feature distillation, how to select valuable features from the dense feature map for effective distillation is also a crucial problem~~\cite{huang, Guo}. Considering that the teacher features are acclimated with the task loss, we can naturally get the importance of features by injecting a attention module between acclimated teacher feature and decoder. The attention module is also optimized with task loss. If a region is important to the performance, its attention weights should be large; in contrast, if a feature is useless to the prediction, we just multiply a zero weight to remove it. Therefore, our proposed loss attention module (LAM), is designed with a cross-attention layer. As illustrated in Figure~\ref{fig:lam}, LAM uses the average of input feature as the query to capture the importance of each pixel, then the output feature is weighted by the importance scores. Besides, the scores, which represents the importances of each pixel, is also returned to compute the distillation loss.

\begin{table*}[t]
\centering
\footnotesize
\renewcommand{\tabcolsep}{1.5mm}

\caption{Performance comparisons on Eigen split od KITTI dataset. All the FLOPs and latencies are measured on a NVIDIA 3090 GPU with $1120\times 352$ input image size.LightDepth have no PyTorch implementation for speedtest.}
\label{tab:kitti}
\resizebox{\textwidth}{!}{
\begin{tabular}{ll|ccc|*{6}{c}c}
\toprule
\textbf{Method} & \textbf{Backbone} &\thead{\textbf{FLOPs}\\\textbf{(B)}} & \thead{\textbf{Params}\\\textbf{(M)}} & \thead{\textbf{Latency}\\\textbf{(ms)}} & \textbf{Abs Rel↓} & \textbf{Sq Rel↓} & \textbf{RMSE↓} & \textbf{log10↓}& \textbf{$\delta_1$↑} & \textbf{$\delta_2$↑} & \textbf{$\delta_3$↑} \\
\midrule
DepthFormer & Swin-T & 1256.6 & 273.7 & 112.1 & 0.052 & 0.158 & 2.143 & 0.079 & 0.975 & 0.997 & 0.999 \\
BinsFormer & Swin-L & 972.7 & 254.6 & 73.0 & 0.079 & 0.151 & 2.092  & 0.079 & 0.974 & 0.997 & 0.999 \\
P3Depth& ResNet101 & 738.6 & 94.3 & 38.9 & 0.071 & 0.270 & 2.842 & 0.103 & 0.953 & 0.993 & 0.999 \\
NeWCRFs & Swin-L & 722.7 & 360.7 & 52.6 & 0.052 & 0.155 & 2.129  & 0.079 & 0.974 & 0.997 & 0.999 \\
PixelFormer & Swin-L & 703.7 & 270.9 & 46.2 & 0.051 & 0.149 & 2.081 & 0.077 & 0.976 & 0.997 & 0.999 \\
AdaBins & MViT & 481.6 & 78.3 & 36.0 & 0.058 & 0.190 & 2.360  & 0.088 & 0.964 & 0.995 & 0.999 \\
BTS & ResNext101 & 474.6 & 112.8 & 28.6 & 0.059 & 0.241 & 2.756 & 0.096 & 0.956 & 0.993 & 0.998 \\
LightDepth & DenseNet196 & 42.8 & 42.6 & --& 0.070 & 0.294 & 2.923  & -- & -- & -- & -- \\
\midrule
DisDepth-Swin-L & Swin-L (teacher) & 800.0 & 237.0 & 47.1 & 0.050 & 0.144 & 2.058  & 0.076 & 0.976 & 0.997 & 0.999 \\
DisDepth-B5 & EfficientNet-B5 & 122.5 & 61.9 & 22.0 & 0.058 & 0.190 & 2.321  & 0.088 & 0.964 & 0.995 & 0.999 \\
DisDepth-B3 & EfficientNet-B3 & 61.1 & 23.1 & 12.2 & 0.060 & 0.201 & 2.366  & 0.096 & 0.957 & 0.995 & 0.999 \\
DisDepth-B0 & EfficientNet-B0 & 35.7 & 8.0 & 6.7 & 0.065 & 0.231 & 2.545 & 0.101 & 0.949 & 0.993 & 0.999 \\
DisDepth-R50 & ResNet-50 & 499.7 & 98.6 & 17.4 & 0.054	&0.171	&2.168 &  0.081 &0.972 &0.997 &0.999 \\
DisDepth-R18 & ResNet-18 & 72.2 & 18.0 & 5.0 & 0.056	&0.184	& 2.298 & 0.086 &0.967	&0.996 & 0.999 \\

\bottomrule
\end{tabular}
}
\end{table*}

\textbf{Attentive distillation loss.} Given the same image as input, we have the output features $\bm{F}_{l}^{(t)}\in\mathbb{R}^{C\times H\times W}$ and $\bm{F}_{l}^{(s)}\in\mathbb{R}^{C\times H\times W}$ of the teacher and student backbones, where $C$, $H$, $W$ denote the numbers of channels, feature height, feature width, respectively, and $l=1,...,L$ is the index of last $L=4$ stages of backbone. Our attentive KD first generates the spatial attentions $\bm{A}_{l}$ using the teacher features, then these attentions representing the importance of each pixel to the depth predictions, are multiplied onto the teacher and student features to guide the student focus on imitating the important regions of teacher features. The attentive KD loss is formulated as
\begin{equation}
    \mathcal{L}_\mathrm{kd} := \sum_{l=1}^{L}||\bm{A}_{l}(\bm{F}^{(s)}_l - \bm{F}^{(t)}_l)||_2^2.
\end{equation}

\textbf{Task loss.} Following previous works~\cite{agarwal}, we use a scaled version of the Scale-Invariant loss (SILog)~\cite{eigen} as our task loss. With the ground-truth depth $d_i^*$ and the predicted depth $\hat{d}_i$ at each pixel index of $n$ pixels, we first calculate the logarithmic distance $g_i = \mathrm{log}(\hat{d}_i) - \mathrm{log}(d^*_i)$, then the SILog is computed as
\begin{equation}
    \mathcal{L}_\mathrm{SILog} = \alpha\sqrt{\frac{1}{n}\sum_{i=1}^n g_i^2 - \frac{\beta}{n^2}\left(\sum_{i=1}^n g_i\right)^2},
\end{equation}
we set $\alpha=10$ and $\lambda=0.85$ in all our experiments.

\textbf{Overall loss.} As a result, our overall loss is the summation of task loss $\mathcal{L}_\mathrm{SILog}$, teacher's task loss $\mathcal{L}_\mathrm{SILog}^{(t)}$, and attentive KD loss $\mathcal{L}_\mathrm{kd}$, \textit{i.e.},
\begin{equation}
    \mathcal{L} = \mathcal{L}_\mathrm{SILog} + \mathcal{L}_\mathrm{SILog}^{(t)} + \lambda \mathcal{L}_\mathrm{kd},
\end{equation}
where $\lambda$ is the factor for balancing loss terms.

\section{Experiments}

\begin{table*}[t]
    \centering
    \footnotesize
    \caption{Performance comparisons on NYU-Depth-v2 dataset. All the FLOPs and latencies are measured on a NVIDIA 3090 GPU with $640\times 480$ input image size.}
    \label{tab:nyu}
    \resizebox{\textwidth}{!}{
    \begin{tabular}{ll|ccc|*{6}{c}c}
        \toprule
        \textbf{Method} & \textbf{Backbone} &\thead{\textbf{FLOPs}\\\textbf{(B)}} & \thead{\textbf{Params}\\\textbf{(M)}} & \thead{\textbf{Latency}\\\textbf{(ms)}} & \textbf{Abs Rel↓} & \textbf{RMSE↓} & \textbf{log10↓} & \textbf{$\delta_1$↑} & \textbf{$\delta_2$↑} & \textbf{$\delta_3$↑} \\
        \midrule
        DepthFormer & Swin-T & 976.7 & 273.7 & 81.5 & 0.096 & 0.339 & 0.041 & 0.921 & 0.989 & 0.998 \\
        BinsFormer & Swin-L & 756.6 & 254.6 & 56.3 & 0.094 & 0.330  & 0.041 & 0.925 & 0.989 & 0.997 \\
        P3Depth & ResNet101 & 575.5 & 94.5 & 29.7 & 0.104 & 0.356 & 0.043 & 0.898 & 0.981 & 0.996 \\
        NeWCRFs  & Swin-L & 561.2 & 270.4 & 40.6 & 0.095 & 0.316 & 0.041 & 0.922 & 0.992 & 0.998 \\
        PixelFormer & Swin-L & 546.2 & 280.3 & 35.0 & 0.090  & 0.322 & 0.039 & 0.929 & 0.991 & 0.998 \\
        AdaBins  & MViT & 375.2 & 78.3 & 24.0 & 0.103 & 0.364 & 0.044 & 0.903 & 0.984 & 0.997 \\
        BTS  & ResNext101 & 369.8 & 112.8 & 22.8 & 0.111 & 0.399 & 0.049 & 0.880 & 0.977 & 0.994 \\
        \midrule
        DisDepth-Swin-L  & Swin-L (teacher) & 621.3 & 237.0 & 35.5 &0.090&0.338	&0.041	& 0.920	&0.991	&0.998 \\
        DisDepth-B5 & Efficientnet-B5 & 95.5 & 61.9 & 17.0 &0.103	&0.369	& 0.044 &0.901	&0.985 &0.997 \\
        DisDepth-B0 & Efficientnet-B0 & 27.8 & 8.0 & 5.3&0.125&0.398	&0.048	& 0.872	&0.977	&0.995 \\

        \bottomrule
    \end{tabular}
    }
\end{table*}
\textbf{Datasets.} Following previous works, we use two popular benchmark datasets KITTI~\cite{Geiger} and NYU Depth V2~\cite{silberman} to validate our efficacy on outdoor and indoor depth estimations, respectively.



\textbf{Models.} The evaluate our method sufficiently, we conduct experiments with various backbone variants, including EfficientNet~\cite{Tan_Le}, GhostNet~\cite{han}, ResNet~\cite{targ}, and Swin Transformer~\cite{liu}.

\textbf{Evaluation Metrics.} We use standard metrics following previous studies~\cite{bhat}, such as average relative error (Abs Rel), root mean squared error (RMSE), and average log error (log10). In addition, we use threshold accuracy ($\delta_i$) at thresholds $\delta$= 1.25, $1.25^2, 1.25^3$ to compare our method to state-of-the-art methods used in earlier works. For KITTI evaluation, we additionally use square relative error (Sq Rel).

\textbf{Training strategies.} The DisDepth is implemented in Pytorch \cite{Paszke}.We use Adam optimizer~\cite{Kingma_Ba_2014} with $\beta_1=0.9$, $\beta_2=0.999$, with a batch size of 8 and weight decay of $10^{-2}$. After a warmup of 7 epochs, distillation begins.We use 25 epoch for both KITTI and NYU-V2 dataset with an initial learning rate of $4 \times 10^{-5}$, which is decreased linearly to $4 \times 10^{-6}$ across the training iterations. We use a similar test protocol as in~\cite{bhat, yuan}. We set KD weight $\lambda=0.05$ in all experiments.

\textbf{Compared methods.} We compare our DisDepth with recent state-of-the-art methods, including NeWCRFs~\cite{yuan}, BTS~\cite{Lee}, P3Depth \cite{Patil}, AdaBins~\cite{bhat}, DepthFormer~\cite{Li_Chen}, BinsFormer~\cite{Li_Zhen}, PixelFormer~\cite{agarwal}, and LightDepth~\cite{Karimi}.

\subsection{Results on KITTI}
On the Eigen split of the KITTI dataset, we compare several state-of-the-art depth estimation methods with our proposed family of DisDepth models, as shown in Table~\ref{tab:kitti}. We can see that, the methods with transformer backbones have a large FLOPs, parameters, and latency, while our models enjoy significant efficiency and performance superiority compared to existing methods. For example, our DisDepth obtains 2.546 RMSE with only 35.7B, surpassing the 42.8B FLOPs LightDepth by a large margin of 0.377. With ResNet-50 backbone, DisDepth-R50 obtains similar performance compared to state-of-the-art approaches PixelFormer, while saving $\sim30\%$ FLOPs.

\subsection{Results on NYU Depth v2}

We summarize the results on NYU Depth v2 dataset in Table~\ref{tab:nyu}. Similar to the results on KITTI, our models achieve competitive performance with obvious reductions on FLOPs, parameters, and latency. For example, our DisDepth-B0 achieves similar performance with only 7.5\% of FLOPs compared to BTS.

\subsection{Ablation Study}

\textbf{Comparisons of student distilled by teacher with or without FA.} In Table~\ref{tab:kd_vs_ours}, we compare our method with the traditional KD method without feature acclimation (FA). The results show that, without feature acclimation, the cross-architecture distillation with Swin-L teacher obtains poor performance compared to EfficientNet-B5 teacher. While our FA can improve the distillation performance of both teachers, and the Swin-L teacher obtains better results since it is more advanced. This implies that the FA is crucial and effective for cross-architecture distillation.

\begin{table}[h]
    \centering
    \footnotesize
    \caption{Comparisons of student distilled by teacher with or without feature acclimation (FA) on the KITTI dataset.}
    \label{tab:kd_vs_ours}
    \begin{tabular}{ll|ccc}
        \toprule
        \textbf{Teacher} & \textbf{Student} & \textbf{Abs Rel↓} & \textbf{Sq Rel↓} & \textbf{RMSE↓}\\
        \midrule
        Eff-B5 & - & 0.0662 & 0.2337 & 2.6020 \\
        Eff-B5 & Eff-B0 & 0.0772 & 0.2896 & 2.7807 \\
        Eff-B5 w/ FA & Eff-B0 & 0.0717 & 0.2599 & 2.6559\\
        \midrule
        Swin-L & - & 0.0503 & 0.1448 & 2.0585 \\
        Swin-L & Eff-B0 & 0.0791 & 0.3320 & 2.9471\\
        Swin-L w/ FA & Eff-B0 & \textbf{0.0650} & \textbf{0.2329} & \textbf{2.5446}\\
        \bottomrule
    \end{tabular}
\end{table}

\textbf{Effects of the proposed components.} We validate the effects of each of our proposed components in Table~\ref{tab:proposed_component_effect}. (1) \textbf{LG.} The student model with our local-global convolution achieves a 0.18 improvement on RMSE. (2) \textbf{KD.} Compared to the independent training, the classical KD method further brings $\sim0.20$ decrease.  (3) \textbf{FAM}. Our feature acclimation significantly improves the performance by 0.27. (4) \textbf{LAM}. Our complete method with loss attention module obtains the optimal performance.

\begin{table}[h]
\centering
\footnotesize
\renewcommand{\tabcolsep}{1.7mm}
\caption{Ablation study of the proposed Disdepth on the KITTI dataset. LG: Local-Global convolution; KD: knowledge distillation; FAM: feature acclimation module; LAM: loss attention module.}
\begin{tabular}{cccc|ccc}
\toprule
\textbf{LG} & \textbf{KD} & \textbf{FAM} & \textbf{LAM} & \textbf{Abs Rel↓} & \textbf{Sq Rel↓} & \textbf{RMSE↓} \\
\midrule
\ & & & & 0.0895 & 0.4271 & 3.3232 \\
\checkmark & & &  & 0.0791 & 0.3896 & 3.1454 \\
\checkmark & \checkmark & & & 0.0743 & 0.3320 & 2.9471 \\
\checkmark & \checkmark & \checkmark & & 0.0721 & 0.2659 & 2.6786 \\
\checkmark & \checkmark & \checkmark & \checkmark & \textbf{0.0650} & \textbf{0.2329} & \textbf{2.5446} \\
\bottomrule
\end{tabular}
\label{tab:proposed_component_effect}
\end{table}

\textbf{Performance of ghost teacher.} To show how the teacher's performance is affected by the ghost decoder, we report the performance of the original teacher and ghost teacher in Table~\ref{tab:ghost_teacher_performance}. We can see that, the ghost decoder only has a minor degradation of the performance. This demonstrate our hypothesis that the task-relevant extrinsic feature can be easily adapted and removing the architecture intrinsic feature would not affect the performance.

\begin{table}[h]
    \centering
    \footnotesize
    \caption{Performance of the teacher with ghost decoder and Swin-L backbone on the KITTI dataset.}
    \begin{tabular}{l|ccc}
        \toprule
        \textbf{Method} & \textbf{Abs Rel↓} & \textbf{Sq Rel↓} & \textbf{RMSE↓} \\ 
        \midrule
        Teacher & 0.050 & 0.1440 & 2.0580 \\
        Teacher w/ ghost decoder&0.053 & 0.1562 & 2.1406 \\
        \bottomrule
    \end{tabular}
    \label{tab:ghost_teacher_performance}
\end{table}

\textbf{Warmup epochs.} Table \ref{tab:warmup_effect} examines the performance of different warmup epochs. We obtain the optimal performance when the number of warmup epochs is set to $7$.

\begin{table}[h]
\centering
\footnotesize
\renewcommand{\tabcolsep}{2.5mm}
\caption{Effect of warmup epochs on the KITTI dataset}
\begin{tabular}{c|ccc}
\toprule
\textbf{Warmup} & \textbf{Abs Rel↓} & \textbf{Sq Rel↓} & \textbf{RMSE↓} \\
\midrule
3  & 0.06876 & 0.2653 & 2.6917 \\
5  & 0.06684 & 0.2394 & 2.6019 \\

7  & \textbf{0.06501} & \textbf{0.2329} & \textbf{2.5446} \\
9  & 0.06692 & 0.2339 & 2.5521 \\
10 & 0.07077 & 0.2562 & 2.6204 \\
\bottomrule
\end{tabular}
\label{tab:warmup_effect}
\end{table}

\textbf{Loss Weight.}
In Table~\ref{tab:lossweight_effect}, we conduct experiments to tune the KD loss weight $\lambda$. We can see that, the $\lambda=0.05$ obtains the optimal performance.

\begin{table}[h]
\centering
\footnotesize
\renewcommand{\tabcolsep}{2.5mm}
\caption{Ablation of loss weight $\lambda$ on KITTI dataset.}
\begin{tabular}{c|ccc}
\toprule
\textbf{Loss Weight} &\textbf{Abs Rel↓}  &\textbf{Sq Rel↓} &\textbf{RMSE↓} \\
\midrule
0.01 & 0.07232 & 0.2641 & 2.6083 \\
0.05 & \textbf{0.06501} & \textbf{0.2329} & \textbf{2.5446} \\
0.1 & 0.06870 & 0.2537 & 2.5960 \\
\bottomrule
\end{tabular}
\label{tab:lossweight_effect}
\end{table}

\textbf{Efficiency comparisons of student with or without local-global convolution.} By incorporating Local-Global Convolutions (LG-Conv) into CNNs, there is a small rise in FLOPs, parameters, and latency, as shown in Table \ref{tab:efficiency}. Nevertheless, the benefit on performance is obvious. This indicates that LG-Conv is an effective tool for enhancing the capabilities of CNNs without imposing significant computational overloads.

\begin{table}[ht]
    \centering
    \footnotesize
    \renewcommand{\tabcolsep}{0.7mm}
    \caption{Efficiency comparisons of student with or without Local-Global Convolutions on KITTI dataset.}
    \label{tab:efficiency}
    \begin{tabular}{l|cccc}
        \toprule
        \textbf{Method} & \textbf{FLOPs (B)} & \textbf{Params (M)} & \textbf{Latency (ms)} & \textbf{RMSE↓}\\
      
        \midrule
        Eff-B0 & 30.4 & 5.9 & 5.2 & 3.3232\\
        Eff-B0 w/ LG & 35.7 & 8.0 & 6.7 & 3.1454\\

        \bottomrule
    \end{tabular}
    
\end{table}

\section{Conclusion}
This work presents DisDepth, an advanced monocular depth estimation (MDE) method that combines the robustness of transformer models with the efficiency of CNN architectures. By incorporating a specially designed local-global convolution module, DisDepth effectively captures both fine-grained details and broader scene context. The introduction of a ghost decoder mechanism streamlines the knowledge transfer process from transformers to CNNs, ensuring digestible knowledge alignment. Additionally, the attentive KD loss focuses on the most valuable features, resulting in more accurate depth predictions. Experimental results on the KITTI and NYU Depth V2 datasets demonstrate DisDepth's superior performance, achieving a harmonious balance between depth accuracy and computational efficiency. DisDepth represents a significant advancement in MDE and holds promise for various applications in computer vision.


\bibliographystyle{ACM-Reference-Format}
\bibliography{sample-base}

\appendix
\section{Comparisons of deploying CNNs and Transformers in Edge Computation}

\subsection{Challenges and Main Difficulties of Deploying Transformers}

Transformers, with their large number of parameters in architectures such as Vision Transformers (ViT)~\cite{Ranftl} and BERT-like~\cite{devlin2018bert} Transformers, present a challenge in edge deployments due to their large model sizes. 
This size, combined with the inherent computational complexity, especially with the self-attention mechanism, demands more power and resources. Such requirements can lead to rapid battery drain, especially in real-time applications such as speech recognition or translation, where latency can be significant on devices with limited processing power. The sequence parallelism inherent in the Transformer architecture aids in highly parallel training. However, efficient deployment of these models is problematic in practice because generative inference progresses one token at a time, and the computation for each token sequentially depends on previously generated tokens~\cite{pope2023efficiently}.
Moreover, while edge deployment frameworks aim to support a broad spectrum of deep learning operations, compatibility issues emerge with specific operations or implementations unique to Transformers. These architectures, which process variable-length sequences, often lead to dynamic computations~\cite{bondarenko2021understanding}. Such computations might not align well with frameworks or devices optimized for static operations.  

Deploying Transformer models on devices such as mobile phones, IoT devices, robots, and especially the Edge TPU presents additional challenges due to their resource constraints. The Edge TPU, in particular, has a limited amount of on-chip memory and computational resources, making it difficult to deploy large Transformer models~\cite{reidy2023efficient}.
Another obstacle is the mandatory conversion of Transformer models to the TensorFlow Lite (TFLite) format to make them compatible with the Edge TPU. This conversion, although crucial, can be tedious and may require additional hardware resources. Furthermore, the TFLite format itself, as the only format supported by the Edge TPU, has its own set of operational limitations, adding complexity to the deployment process.

\subsection{Advantages of CNNs over Transformers}

Convolutional neural networks (CNNs), including architectures such as VGG\cite{simonyan2014very} and ResNet\cite{targ}, are characterised by their computational simplicity, using operations such as  convolutions and pooling. In general, these models are smaller, with inventions such as MobileNet~\cite{howard2017mobilenets} and EfficientNet~\cite{Tan_Le} that have specifically been optimised for edge devices. The long history of CNNs in this areas shows that tools including TensorRT, ONNX, and TensorFlow Lite have been made available with extensive optimisation techniques. Importantly, modern Graphical Processing Units (GPUs) and Application-Specific Integrated Circuits (ASICs) (such as Tensor Processing Units (TPUs) and Neural Processing Units (NPUs)) provide dedicated acceleration for convolution operations, thus making CNNs particularly efficient for deployment in real-world scenarios.

\end{document}